\documentclass[11pt]{article}

\usepackage[final]{acl}

\usepackage{times}
\usepackage{latexsym}

\usepackage[T1]{fontenc}

\usepackage[utf8]{inputenc}

\usepackage{microtype}

\usepackage{inconsolata}

\usepackage{graphicx}
\usepackage{algorithm}
\usepackage{algorithmic}
\usepackage{amsmath}
\usepackage{booktabs}
\usepackage{makecell}

\title{Confident Rankings with Fewer Items: Adaptive LLM Evaluation with Continuous Scores}

\author{
  \textbf{Esma Balk{\i}r\textsuperscript{1}},
  \textbf{Alice Pernthaller\textsuperscript{1}},
  \textbf{Marco Basaldella\thanks{Work done while at Trismik.}}, \\
  \textbf{José Hernández-Orallo\textsuperscript{2,3}\thanks{Joint senior authors.}},
  \textbf{Nigel Collier\textsuperscript{1,4}\footnotemark[2]}
\\
\\
  \textsuperscript{1}Trismik,
  \textsuperscript{2}Leverhulme Centre for the Future of Intelligence, University of Cambridge,\\
  \textsuperscript{3}Universitat Politècnica de València,
  \textsuperscript{4}University of Cambridge
\\
  \texttt{\{esma, alice\}@trismik.com, \{jh2135, nhc30\}@cam.ac.uk}
  }

\begin{document}
\maketitle
\begin{abstract}
Computerized Adaptive Testing (CAT) has proven effective for efficient LLM
evaluation on multiple-choice benchmarks, but modern LLM evaluation
increasingly relies on generation tasks where outputs are scored continuously
rather than marked correct/incorrect. We present a principled extension of
IRT-based adaptive testing to continuous bounded scores (ROUGE, BLEU,
LLM-as-a-Judge) by replacing the Bernoulli response distribution with a
heteroskedastic normal distribution.
Building on this, we introduce an uncertainty aware ranker with adaptive stopping criteria that achieves reliable model ranking while testing as few items and as cheaply as possible. We validate our method on five benchmarks spanning n-gram-based, embedding-based, and LLM-as-judge metrics. Our method improves ranking correlation by 0.13 $\tau$ over random sampling and has 99\% accuracy on confident predictions while using 2\% of the items after a one-time calibration step.
\end{abstract}

\section{Introduction}

Rigorous evaluation of large language models is essential, but current
practice faces many challenges. One is cost: exhaustive evaluation
becomes expensive as the number of models, test items, and metrics grows.
Another is methodological: score differences are often reported without
significance testing, leading many studies to mistake statistical noise with improved performance and produce non-replicable results  \citep{dror2018hitchhikers}.
Efficient evaluation methods that maintain statistical validity remain underexplored.

Computerized Adaptive Testing (CAT) and Item Response Theory (IRT) originate in psychometrics, where they estimate test-takers' latent ability on a construct of interest. In ML evaluation \cite{liu2024survey}, the same statistical machinery has been adopted as scaffolding for efficient sampling: identifying representative subsets or adaptively selecting items so that rankings can be recovered with far fewer evaluations than exhaustive benchmarking \citep{martinez2016making, polo2024tinybenchmarks, rodriguez2021evaluation, kipnis2025metabench}, with recent work extending this to fully adaptive item selection \citep{hofmann2025fluid}. We follow this efficiency-focused lineage: we assume that the benchmark scores reflect a capability that the practitioner is interested in, and aim to recover the ranking that full evaluation would produce.

Existing CAT approaches for LLM evaluation focus exclusively on multiple-choice datasets where responses can only be correct or incorrect. However, many of the tasks that reflect real use cases of LLMs such as summarization, dialogue, instruction following and machine translation must be scored on a continuous scale. These scoring strategies cover traditional metrics such as BLEU \citep{papineni2002bleu} and ROUGE \citep{lin2004rouge}, and embedding similarity-based scores such as BERTScore \citep{zhang2020bertscore} and COMET \citep{rei2020comet}. LLM-as-judge evaluation  \citep{zheng2024judging, liu2023gpteval}, now widely adopted for assessing instruction following and open-ended generation, also produces ordinal ratings that can be normalized and treated as continuous.

We present a continuous extension of IRT-based adaptive testing that accommodates real-valued scores while preserving the mathematical structure of binary CAT. Our key insight is to replace the Bernoulli response distribution with a heteroskedastic normal distribution that maintains the same logistic mean function while introducing variance that mimics the Bernoulli structure. This preserves the natural property that variance is highest when the expected score $\mu$ is most uncertain ($\mu = 0.5$) and shrinks at the boundaries. 


Continuous CAT already reduces evaluation cost substantially, but further
savings are possible when we consider how evaluation results are typically used.
Model evaluation is fundamentally comparative: the primary goal is often to determine whether one model is better than others on a given benchmark. Building on continuous CAT, we introduce an adaptive multi-model ranking method that efficiently obtains statistically significant rankings by monitoring uncertainty estimates around model scores. Rather than testing each model to a fixed precision threshold, we stop as soon as models are well-differentiated based on pairwise confidence at a user-specified level. This focuses the testing effort on close competitors while reducing items for clearly separated models. Within this framework, we further observe that when distinguishing two models, testing either one yields useful information, allowing cost-aware allocation that preferentially tests cheaper models to maximize uncertainty reduction per dollar spent.

We validate our approach on five generation benchmarks that span summarization (GovReport, BioLaySumm2025), named entity recognition (Nemotron PII), question answering (TruthfulQA) and translation (FLORES). For each dataset, we evaluate 5 disjoint sets of 4 hold-out models and multiple metric types (ROUGE, BLEU, BERTScore, COMET, readability indices, LLM-as-judge). Our adaptive ranker achieves 0.75 Kendall's $\tau$ correlation with ground-truth rankings while using only 2\% of the full evaluation budget at evaluation time, outperforming random sampling by 0.13 $\tau$. This figure excludes the one-time cost of calibrating item parameters, which is amortized over subsequent evaluations and breaks even after roughly 70 model evaluations on a dataset (Appendix~\ref{app:cost}). Compared to fixed-length CAT, adaptive stopping provides an additional 32\% item reduction and 39\% cost savings.

Our contributions are:
\begin{enumerate}
    \item A principled extension of IRT-based CAT from binary to continuous bounded outcomes via heteroskedastic normal distribution
    \item Adaptive multi-model ranking algorithm with pairwise stopping and cost-aware allocation
    \item Empirical validation across diverse generation tasks, metrics, and model scales
\end{enumerate}

The continuous CAT framework opens adaptive testing to the full spectrum of modern LLM evaluation, and the adaptive multi-model ranker enables efficient comparisons of generation quality across model candidates. We make our code available.\footnote{\url{https://github.com/trismik/continuous-cat}}

\section{Related Work}

Early studies on efficient model evaluation used IRT to analyze benchmark properties such as item difficulty and discrimination \citep{martinez2016making,lalor2016building, vania2021comparing, rodriguez2021evaluation}.
\citet{polo2024tinybenchmarks} and \citet{kipnis2025metabench} used IRT to identify representative static subsets, while \citet{hofmann2025fluid} developed a fully adaptive framework that dynamically selects items during evaluation. However, all existing approaches assume binary outcomes and cannot accommodate the continuous scores.  \citet{lalor2016building} applied IRT to construct evaluation scales for NLP systems, and \citet{prudencio2015analysis} and \citet{lalor2019learning} showed that IRT models can be fit using response patterns from model ensembles rather than human annotations.
\citet{chen2019beta} proposed a Beta-distributed IRT model applied to modelling classifier confidence scores.

This follows the tradition in psychometrics, with \citet{samejima1973homogeneous} introducing the continuous response model, and \citet{noel2007beta} proposing Beta-distributed models that naturally respect $[0,1]$ bounds. Both introduce additional complexity to Fisher Information calculations, complicating item selection.  Beta models also cannot accommodate exact boundary values, requiring either ad hoc transformations or zero-and-one inflated extensions \citep{molenaar2022zero}. Related continuous-response formulations include linear factor-analytic indices for difficulty and information, and Rasch-type models for continuous responses in large-scale learning systems \citep{ferrando2009difficulty,deonovic2020rasch}. These typically assume homoskedastic residuals or prioritize scalable scoring over closed-form CAT information. \citet{mellenbergh1994generalized} developed generalized linear IRT with simpler closed-form solutions, but assumes constant variance across the score range. Our heteroskedastic normal formulation combines simplicity with appropriate variance structure and maintains direct compatibility with standard CAT algorithms.

\section{Background}

\paragraph{Item Response Theory (IRT)} \citep{hambleton1991fundamentals} models the probability of a correct response as a function of the test-taker's latent ability $\theta$ and item parameters. The simplest model, the 1-Parameter Logistic (1PL) model, defines this probability as:

\begin{equation}  \label{eq:1pl}
P(X=1|\theta, a, b) = \frac{1}{1 + \exp(-a(\theta - b))}
\end{equation}

where $b$ is item difficulty and $a$ is a discrimination parameter shared across all items. When ability matches difficulty ($\theta = b$), the probability of success is exactly 0.5. Higher $a$ yields steeper response curves that better differentiate between ability levels.

 \begin{figure}[t]
  \centering
  \includegraphics[width=\columnwidth]{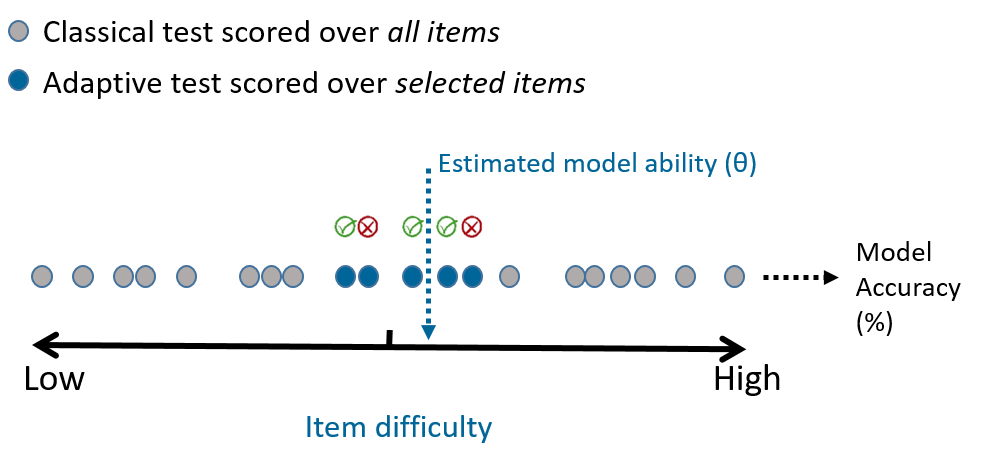}
  \caption{Adaptive testing focuses on items around the model ability, skipping those for which it would most certainly (i.e., uninformatively) get high or low scores.}
  \label{fig:adaptive-testing}
  \end{figure}

\paragraph{Computerized adaptive testing (CAT)} \citep{wainer2000computerized} dynamically selects items to efficiently estimate a test-taker's ability (see Figure~\ref{fig:adaptive-testing}). The algorithm initializes with a prior distribution over $\theta$, typically $\mathcal{N}(0, \sigma_0^2)$ where the prior variance is a hyperparameter controlling initial uncertainty. At each iteration, it selects the item that maximizes Fisher Information at the current ability estimate
$i^* = \arg\max_i I(\hat{\theta}|b_i)$. After observing the response, the ability estimate and posterior are updated via Bayesian updating. As items accumulate, the standard error
$
\text{SE}(\hat{\theta}) = 1/\sqrt{\sum_{i=1}^{n} I_i(\hat{\theta})}
$ decreases.
Testing terminates when it falls below a predetermined threshold.

\paragraph{Fisher Information} quantifies how much an observation tells us about $\theta$. For the 1PL model with binary outcomes, this is derived directly from Bernoulli variance:
\begin{equation}
I(\theta|a, b) = a^2 \cdot P(\theta, b) \cdot (1 - P(\theta, b))
\end{equation}
Information is highest when $P = 0.5$ (maximum uncertainty about the outcome) and approaches zero as $P$ approaches 0 or 1 (outcome nearly certain). This creates heteroskedastic variance as a function of the predicted probability $P(\theta, b)$.
\begin{figure*}[t]
  \centering
  \includegraphics[width=\textwidth]{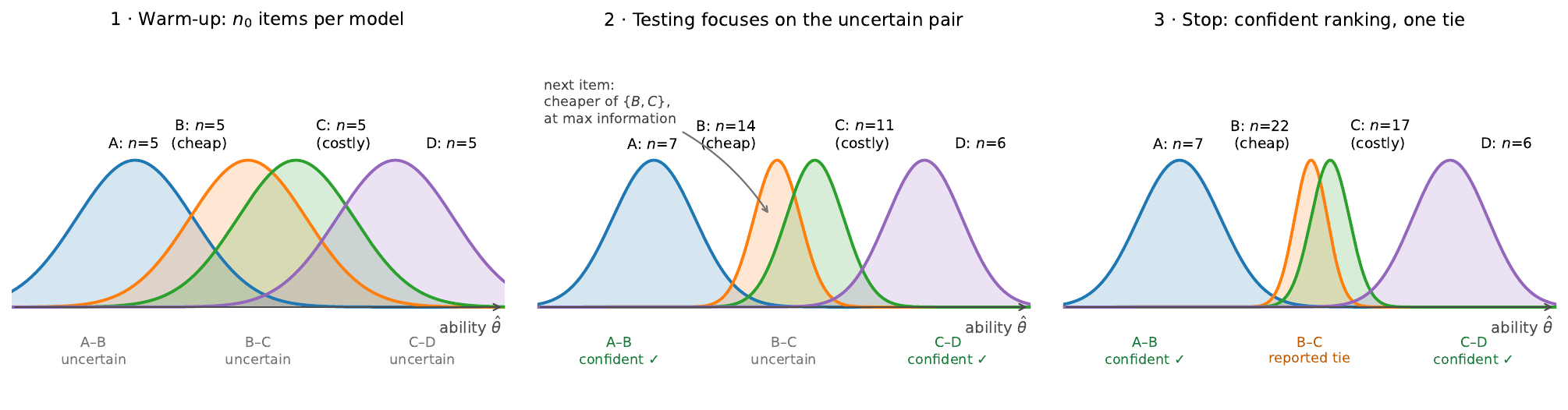}
  \caption{The adaptive multi-model ranker on four models $A$--$D$ (illustrative). After warm-up, posteriors over ability $\theta$ are wide and every adjacent pair in the ranking is uncertain (left). Testing then concentrates on the remaining uncertain pair: clearly separated models stop receiving items, and the next item goes to the cheaper model of the pair, chosen at maximum Fisher information (middle). The run stops when every adjacent pair is either confidently ordered or reported as a tie at the budget (right). $n$ is the number of items administered to each model.}
  \label{fig:ranker-schematic}
\end{figure*}

\section{Continuous CAT}
\label{sec:continuous-cat}

To extend CAT to continuous outcomes, we require a response distribution for scores in $[0,1]$ rather than binary $\{0,1\}$. We retain the logistic mean function from standard IRT, which captures the relationship between ability, difficulty, and expected performance, and replace the Bernoulli distribution with a normal distribution:

\begin{equation}
X|\theta, b, k \sim \mathcal{N}(\mu(\theta, b), \sigma^2(\theta, b))
\end{equation}
where the mean follows the logistic function:
\begin{equation}
\mu(\theta, b) = \frac{1}{1 + \exp(-(\theta - b))}
\end{equation}
We preserve the Bernoulli variance structure $\text{Var}(X) = P(1-P)$ by defining:
\begin{equation}
\sigma^2(\theta, b) = k \cdot \mu(\theta, b) \cdot (1 - \mu(\theta, b))
\end{equation}
where $k = 1/a^2$ is a noise parameter capturing measurement precision. We estimate a single $k$ per dataset-metric combination from calibration data.

Since both the mean and the variance of the response distribution depend on $\theta$, the Fisher Information for this model has two terms:

\begin{equation} \label{eq:full-fi}
I(\theta|b, k) = \frac{\mu(1-\mu)}{k} + \frac{(1-2\mu)^2}{2}
\end{equation}

Throughout, we use the first term $\tilde{I}(\theta|b, k) = \mu(1-\mu)/k$ and ignore the second term.
This simplification is safe since the full information is linear in $u = \mu(1-\mu)$, so it orders items identically to $\tilde{I} \propto u$ whenever $a = 1/\sqrt{k} > \sqrt{2}$ (8 of our 9 dataset-metric pairs), and the omitted term is nonnegative, so $\tilde{I}$ only overstates standard errors. Appendix~\ref{app:full-fi} gives the derivation and an empirical comparison confirming that the full FI changes nothing on the $a > \sqrt{2}$ pairs and destabilizes the one low-discrimination dataset.

 Comparing $\tilde{I}$ with the 1PL Fisher Information $I = a^2 \cdot P(1-P)$, we see that $k = 1/a^2$: high discrimination ($a > 1$) corresponds to low noise ($k < 1$), and vice versa. Both parameterizations capture the same underlying question: how reliably does an observed response reflect $\theta$? This equivalence indicates that our continuous extension constitutes a natural generalization that unifies the treatment of measurement precision across binary and continuous settings.

A limitation of the normal distribution is its unbounded support: it can technically assign probability to values outside $[0,1]$. However, the heteroskedastic variance structure mitigates this concern: as $\mu$ approaches the boundaries, $\sigma^2 = k \cdot \mu(1-\mu)$ shrinks, concentrating the distribution within the valid range. Alternatives such as the $\beta$-distribution are naturally bounded but do not preserve the connection to binary IRT and complicate the Fisher Information derivation. In practice, we find that the normal approximation performs well across the metrics we consider.

  The CAT algorithm itself remains unchanged. Item selection still maximizes Fisher Information at the current ability estimate, with the information formula now
  incorporating $k$. Posterior updates follow the same Bayesian logic with normal likelihoods. The standard error formula $\text{SE}(\hat{\theta}) = 1/\sqrt{\sum_i I_i}$
  applies as before.

  \section{Parameter Estimation}
  \label{sec:parameter-estimation}

 We estimate both $b_i$ and $k$ from historical evaluation data by fitting the model to observed score distributions across a population of models.

  \paragraph{Item Difficulty.} We adopt the 1PL model and estimate item difficulties from observed scores across calibration models. Following standard IRT convention, we center the ability scale such that the mean
  model ability is zero. Under this parameterization, the maximum likelihood estimate for item difficulty simplifies to:

  \begin{equation}
  b_i = -\text{logit}(\hat{p}_i) = \log\left(\frac{1 - \hat{p}_i}{\hat{p}_i}\right)
  \end{equation}
 where $\hat{p}_i$ is the average score on item $i$ across all calibration model-temperature configurations. When scores cluster in a narrow range (e.g. like ROUGE where most models achieve 0.3--0.5), the resulting
   difficulties may not span the full ability range. We therefore apply min-max normalization to $[\epsilon, 1-\epsilon]$ before the logit transformation.

\paragraph{Noise Parameter.} We estimate a global $k$ from the calibration data using method-of-moments. Given item difficulties $\{b_i\}$ and model ability estimates
  $\{\theta_j\}$ (computed as $\theta_j = \text{logit}(\bar{y}_j)$ where $\bar{y}_j$ is the model's average scaled score), we obtain:

  \begin{equation}
  k = \frac{\sum_{i,j} (y_{ij} - \mu_{ij})^2}{\sum_{i,j} \mu_{ij}(1-\mu_{ij})}
  \end{equation}

  where $\mu_{ij} = \text{logit}^{-1}(\theta_j - b_i)$ is the predicted mean score for model $j$ on item $i$. This formula directly estimates the variance inflation factor:
   when $k=1$, observed variance matches the Bernoulli structure exactly; $k>1$ indicates noisier measurements than binary responses; $k<1$ indicates more precise
  measurements.

\paragraph{Filtering Negative Discrimination Items.}
  Some items may anti-correlate with model ability. On these items, weaker models score higher than stronger models, rendering them effectively uninformative. Because we use a 1PL model with a global discrimination parameter, we filter such items by computing the Pearson correlation between item scores and model abilities across the calibration set, and excluding items with negative correlation from the adaptive item pool.

\section{Adaptive Ranker}

\begin{algorithm*}[t]
\caption{Adaptive Multi-Model Ranking}
\label{alg:adaptive-ranker}
\begin{algorithmic}[1]
\REQUIRE Models $\{1, \ldots, M\}$, item bank $\{b_i\}$ with global noise $k$, costs $\{c_m\}$, confidence $\gamma$, max items $n_{\text{max}}$
\STATE Initialize $\hat{\theta}_m \sim \mathcal{N}(\text{median}({b}_i), 25)$,
\STATE \textbf{Warm-up:} Administer $n_{\text{init}}$ items to each model using MFI selection
\WHILE{not all adjacent pairs confident or at max items}
    \STATE Rank models by $\hat{\theta}_m$
    \STATE Identify uncertain pairs: $\mathcal{U} = \{(i,j) : \gamma > P(\theta_i > \theta_j) > 1-\gamma \text{ and } n_i < n_{\text{max}} \text{ and } n_j < n_{\text{max}}\}$
    \IF{$\mathcal{U} = \emptyset$}
        \STATE \textbf{break}
    \ENDIF
    \STATE Collect candidate models: $\mathcal{C} = \{m : m \in (i,j) \text{ for some } (i,j) \in \mathcal{U}\}$
    \STATE Select model: $m^* = \arg\max_{m \in \mathcal{C}} \frac{\text{SE}_m^2}{(n_m + 1) \cdot c_m}$
    \STATE Select item: $i^* = \arg\max_i I(\hat{\theta}_{m^*} | b_i, k)$
    \STATE Administer item $i^*$ to model $m^*$, observe score $y$
    \STATE Update $\hat{\theta}_{m^*}$ and $\text{SE}_{m^*}$
\ENDWHILE
\RETURN Ranking with pairwise confidence scores
\end{algorithmic}
\end{algorithm*}


Standard CAT runs each model independently until its own standard error falls below the fixed precision threshold $\epsilon$ introduced in \S\ref{sec:continuous-cat}. However, when model ranking is the main goal and $\hat{\theta}$ values are an intermediate quantity used to place all models on a shared scale, this is inefficient: a model whose ability lies several standard errors above the next-best receives the same item budget as two models within half a standard error of each other, even though only the latter pair requires further testing to establish their relative order. We propose an adaptive multi-model ranking algorithm (Algorithm~\ref{alg:adaptive-ranker}) that directly optimizes for the comparative goal. Given $M$ models to evaluate on a shared item bank, the algorithm runs independent CAT estimators for each model, but monitors pairwise confidence in the ranking and terminates when all adjacent model pairs are statistically well-differentiated at a user-specified confidence level. Figure~\ref{fig:ranker-schematic} illustrates the procedure.

\paragraph{Pairwise confidence.} Given ability estimates $\hat{\theta}_i$ and $\hat{\theta}_j$ with standard errors $\text{SE}_i$ and $\text{SE}_j$, we compute the probability that model $i$ outperforms model $j$ under a normal approximation:

\begin{equation}
P(\theta_i > \theta_j) = \Phi\left(\frac{\hat{\theta}_i - \hat{\theta}_j}{\sqrt{\text{SE}_i^2 + \text{SE}_j^2}}\right)
\end{equation}

where $\Phi$ is the standard normal CDF. This quantifies our confidence in the pairwise ordering given current uncertainty in both estimates.

\paragraph{Adaptive stopping.} Traditional CAT terminates when each model's standard error falls below a threshold $\epsilon$, regardless of how the models compare. We instead terminate when all adjacent pairs in the current ranking satisfy $P(\theta_i > \theta_j) > 1 - \frac{1-\gamma}{2}$ or $P(\theta_i > \theta_j) < \frac{1-\gamma}{2}$, corresponding to a two-sided confidence
  interval at level $\gamma$, or we reach a maximum item budget $n_{\text{max}}$. This focuses testing effort where it matters: models that are clearly separated require fewer items to establish their relative ordering, while close competitors receive additional testing until their difference reaches statistical significance or the budget is exhausted. If two models remain indistinguishable at the budget limit, they can be reported as tied rather than forced into an arbitrary ordering.

\paragraph{Cost-aware allocation.} When a pair $(i, j)$ requires additional testing to reach confidence threshold $\gamma$, we must choose which model to evaluate next. Since testing either model reduces uncertainty in the pairwise comparison, we select the model that provides the greatest expected uncertainty reduction per unit cost:

\begin{equation}
\text{value}_m = \frac{\text{SE}_m^2}{(n_m + 1) \cdot c_m}
\end{equation}
where $n_m$ is the number of items already administered to model $m$ and $c_m$ is its per-item evaluation cost. The numerator reflects current uncertainty; the denominator captures diminishing returns (each additional item contributes less) weighted by cost. This criterion naturally allocates more items to cheaper models when resolving uncertain comparisons, yielding additional cost savings beyond item reduction alone.

\paragraph{Algorithm.} The complete procedure combines continuous CAT estimation with pairwise stopping and cost-aware allocation under a total budget constraint. Rather
  than allocating a fixed maximum number of items per model, we specify a total cost budget $B$ that can be flexibly distributed across models. After a warm-up phase that
  administers initial items to all models via MFI selection, the algorithm iteratively identifies uncertain pairs, selects the most cost-effective model to test, chooses
  the maximally informative item for that model, and updates posteriors. Testing continues until either all adjacent pairs reach the confidence threshold, or the total budget $B$ is exhausted.

\section{Experiments}

We evaluate our continuous CAT extension and adaptive ranker across five generation tasks spanning summarization, machine translation, question answering, and named entity recognition. Our goal is to approximate ground-truth rankings from full dataset evaluation using as few items as possible.

\subsection{Experimental Protocol}
\label{sec:protocol}

We select tasks representing diverse generation settings and metric types (Table~\ref{tab:datasets}). This selection covers n-gram overlap (ROUGE, BLEU), learned embeddings (BERTScore, COMET), readability indices (FKGL), LLM-as-judge, and span-level F1. We exclude BERTScore from GovReport as all models achieved near-identical scores (range $<$0.02), yielding no meaningful ground-truth ranking. Full prompts and scoring details are provided in Appendix~A.

\begin{table*}[t]
\centering
\small
\caption{Evaluation datasets. FKGL: Flesch-Kincaid Grade Level \citep{kincaid1975derivation}.}
\label{tab:datasets}
\renewcommand{\arraystretch}{1.2}
\begin{tabular}{lrcll}
\toprule
\textbf{Dataset} & \textbf{Size} & \textbf{Split} & \textbf{Task} & \textbf{Metrics} \\
\midrule[0.08em]
\makecell[l]{BioLaySumm \citep{goldsack2024biolaysumm}} & 1,376 & val & Lay summarization & \makecell[l]{ROUGE-L, BERTScore, FKGL} \\
\midrule
\makecell[l]{GovReport \citep{huang2021efficient}} & 973 & val & Document summarization & ROUGE-L \\
\midrule
\makecell[l]{TruthfulQA \citep{lin2022truthfulqa}} & 817 & val & Open-ended QA & \makecell[l]{LLM-Judge, BERTScore (diff.)} \\
\midrule
\makecell[l]{FLORES \citep{nllb2022}} & 1,012 & devtest & Translation (TR$\rightarrow$EN) & \makecell[l]{COMET, BLEU} \\
\midrule
\makecell[l]{Nemotron-PII \citep{nemotron-pii-2025}} & 2,000 & test\footnote{Randomly sampled from full test set.} & NER/PII detection & F1 (span-level) \\
\bottomrule
\end{tabular}
\end{table*}

We evaluate 21 models from 6 families: OpenAI GPT, Google Gemini, Amazon Nova, Meta Llama, Mistral, and Qwen (see Appendix~A for full list). For each dataset-metric pair, we collect exhaustive evaluation scores by running all models on all items, establishing ground-truth rankings. Each model is evaluated at four temperature settings ($T \in \{0.0, 0.4, 0.7, 1.0\}$), which yields 84 model-temperature configurations per item. Our main experiment uses five-fold cross-validation over models. In each fold, we randomly select 4 models as the hold-out set to be ranked adaptively; all temperature variants of hold-out models are excluded from calibration. The remaining 17 models (68 configurations) provide calibration data for estimating item parameters $(b_i, k)$. We fit parameters via maximum likelihood on the calibration set, then simulate adaptive evaluation on the hold-out models at $T=0$. We set the confidence threshold $\gamma = 0.95$ for pairwise decisions and require a minimum of 10 items per model before stopping.

For each hold-out set, we compare the adaptive ranker to two baselines at the same matched cost. \emph{Random sampling} allocates the total budget $B$ across models by iteratively selecting a random affordable model and a random unseen item to administer, stopping at the first draw whose cost exceeds the remaining budget. A \emph{most-informative static subset} transposes static subset selection \citep{polo2024tinybenchmarks, kipnis2025metabench} to continuous scores: per fold, items are ranked by their corrected item--total correlation on the calibration models and accumulated greedily until the next item would exceed the budget; the identical subset is administered to every holdout model, which is then ranked by its mean normalized score. We assess ranking quality using Kendall's $\tau$ correlation between predicted and ground-truth rankings computed from point estimates (mean $\hat{\theta}$). For efficiency, we report the mean number of items administered per model.

\paragraph{Budget.} We set $B$ to the cost equivalent of 20 items per model. At this budget the adaptive ranker is in a budget-constrained regime: its pairwise-confidence stopping rule would naturally fire at roughly 30--100 items per model depending on dataset. Appendix~\ref{app:budget-sweep} traces ranking quality as a function of $B$ and shows that adaptive outperforms random sampling across the small-to-mid budget regime that motivates this work, with the gap shrinking as $B$ approaches full evaluation.


\subsection{Main Results}
\label{sec:main-results}

\begin{table*}[t]
\centering
\small
\caption{Ranking performance across benchmarks. Higher discrimination ($a$, mean over calibration folds) indicates more informative items. Items is the total number of items administered across the four hold-out models. \% Used is Items divided by ($4 \times$ Size), i.e., the fraction of the full hold-out evaluation. Random and Static are the two matched-cost baselines of \S\ref{sec:protocol}; the static subset (15--21 items) is deterministic within a fold, so its $\tau$ is aggregated over the 5 folds.}
\label{tab:main-results}
\begin{tabular}{llccccccc}
\toprule
Dataset & Metric & Size & $a$ & Adaptive $\tau$ $\uparrow$ & Random $\tau$ $\uparrow$ & Static $\tau$ $\uparrow$ & Items & \% Used $\downarrow$ \\
\midrule
BioLaySumm & ROUGE-L & 1376 & 4.10 & \textbf{0.957}$\pm$0.11 & 0.853$\pm$0.20 & 0.800$\pm$0.16 & 81$\pm$18 & 1.5$\pm$0.3\% \\
BioLaySumm & BERTScore & 1376 & 3.57 & \textbf{0.930}$\pm$0.14 & 0.750$\pm$0.36 & 0.533$\pm$0.45 & 90$\pm$10 & 1.6$\pm$0.2\% \\
BioLaySumm & FKGL & 1376 & 2.42 & \textbf{0.777}$\pm$0.28 & 0.750$\pm$0.38 & 0.733$\pm$0.39 & 91$\pm$15 & 1.6$\pm$0.3\% \\
GovReport & ROUGE-L & 973 & 4.67 & 0.813$\pm$0.18 & \textbf{0.820}$\pm$0.24 & 0.667$\pm$0.21 & 95$\pm$29 & 2.4$\pm$0.8\% \\
TruthfulQA & LLM-Judge & 817 & 2.63 & \textbf{0.637}$\pm$0.39 & 0.417$\pm$0.43 & 0.600$\pm$0.25 & 93$\pm$12 & 2.8$\pm$0.4\% \\
TruthfulQA & BERTScore & 817 & 2.68 & 0.520$\pm$0.38 & 0.193$\pm$0.46 & \textbf{0.533}$\pm$0.45 & 93$\pm$7 & 2.9$\pm$0.2\% \\
FLORES & BLEU & 1012 & 3.15 & 0.760$\pm$0.23 & 0.563$\pm$0.43 & \textbf{0.933}$\pm$0.13 & 91$\pm$8 & 2.2$\pm$0.2\% \\
FLORES & COMET & 1012 & 3.96 & 0.670$\pm$0.28 & 0.513$\pm$0.41 & \textbf{0.800}$\pm$0.27 & 101$\pm$10 & 2.5$\pm$0.3\% \\
Nemotron & F1 & 2000 & 1.39 & 0.673$\pm$0.27 & 0.717$\pm$0.28 & \textbf{0.800}$\pm$0.16 & 52$\pm$4 & 0.6$\pm$0.1\% \\
\midrule
\multicolumn{2}{l}{\textit{Overall (mean)}} & & 3.17 & \textbf{0.75} & 0.62 & 0.71 & 87 & 2.0\% \\
\bottomrule
\end{tabular}
\end{table*}

Table~\ref{tab:main-results} presents results across all dataset-metric combinations.
On 7 of 9 dataset-metric pairs, the adaptive ranker achieves higher Kendall's $\tau$ than the random sampling baseline. The largest gains occur on discriminative metrics: +0.10 $\tau$ on BioLaySumm ROUGE-L (0.957 vs 0.853), +0.33 on TruthfulQA BERTScore (0.520 vs 0.193), and +0.20 on FLORES BLEU (0.760 vs 0.563). These results demonstrate that our adaptive ranker provides genuine performance gains over random sampling. The adaptive ranker uses 52--101 items per hold-out set, and achieves 0.75 $\tau$ with the full dataset ranking while using only 2\% of the items per new-model evaluation (excluding the one-time calibration cost; Appendix~\ref{app:cost}).

The discrimination parameter $a$ predicts where adaptive methods provide the most benefit. On high-discrimination metrics ($a > 3$), the adaptive ranker matches or outperforms random sampling. On low-discrimination metrics like Nemotron F1 ($a = 1.39$), individual items provide less information, limiting the advantage of intelligent selection.

\paragraph{Comparison to the informed static subset.} Although our model cannot target discrimination explicitly, the items it selects are typically also the items whose scores vary most across models, so adaptive selection could still act as an implicit proxy for concentrating evaluation on empirically discriminative items. We test whether this is the main source of the adaptive method's gains by administering a static highest discrimination set with matched cost. We find that the static subset trails adaptive testing on average ($\tau = 0.71$ vs 0.75), although it wins on 4 of 9 pairs, against 4 for adaptive and one for random. However, a fixed subset yields only a point ranking, identical whether models are cleanly separated or statistically indistinguishable, whereas the adaptive ranker additionally provides pairwise confidence, early stopping, cost-aware allocation, and tie detection. The out-of-family evaluation (\S\ref{sec:family-holdout}) also shows that the static subset method is more brittle when ranking models less similar to those found in the calibration set.

 \begin{table*}[t]
  \centering
  \small
  \caption{Tie detection against ground-truth ties from bootstrap CIs on full evaluation. Adapt/GT Tie\% are the fraction of pairs reported as ties for adaptive ranker and ground truth. Tie P/R/F1 are the precision, recall and F1 on tie detection quality of the adaptive ranker compared to the full evaluation. Conf Acc is the accuracy of the adaptive ranker on its non-tie predictions.}
  \label{tab:tie-detection}
  \begin{tabular}{llcccccc}
  \toprule
  Dataset & Metric & Adapt Tie\% & GT Tie\% & Tie P $\uparrow$ & Tie R $\uparrow$ & Tie F1 $\uparrow$ & Conf Acc $\uparrow$ \\
  \midrule
  BioLaySumm & ROUGE-L & 13\% & 3\% & 0.63 & 0.99 & 0.66 & 1.00 \\
  BioLaySumm & BERTScore & 40\% & 13\% & 0.32 & 0.97 & 0.39 & 1.00 \\
  BioLaySumm & FKGL & 37\% & 17\% & 0.39 & 0.92 & 0.45 & 0.98 \\
  GovReport & ROUGE-L & 18\% & 10\% & 0.42 & 0.76 & 0.47 & 0.94 \\
  TruthfulQA & LLM-Judge & 79\% & 47\% & 0.52 & 0.91 & 0.61 & 0.98 \\
  TruthfulQA & BERTScore & 98\% & 73\% & 0.74 & 1.00 & 0.82 & 1.00 \\
  FLORES & BLEU & 59\% & 30\% & 0.44 & 1.00 & 0.55 & 1.00 \\
  FLORES & COMET & 54\% & 37\% & 0.52 & 0.92 & 0.54 & 0.98 \\
  Nemotron & F1 & 61\% & 7\% & 0.09 & 1.00 & 0.14 & 0.98 \\
  \midrule
  \multicolumn{2}{l}{\textit{Overall (mean)}} & 51\% & 26\% & 0.45 & 0.94 & 0.51 & 0.99 \\
  \bottomrule
  \end{tabular}
  \end{table*}

\paragraph{Distributional Conformance.} In practice, metrics conform to the assumed variance structure $\sigma^2 = k \cdot \mu(1-\mu)$ to widely varying degrees, yet ranking quality does not track this conformance: the best-ranking metric is among the worst conformers, and it is low discrimination rather than poor fit that marks the settings where adaptive selection struggles. Appendix~\ref{app:conformance} presents the analysis; violations of the variance assumption do not degrade ranking performance when items are sufficiently discriminative.

\begin{table*}[t]
\centering
\small
\caption{Isolating the stopping rule. Fixed-length CAT administers $n = \max$ items adaptive used per holdout to every model, so it sees a strict superset of adaptive's items in the same holdout. Any $\tau$ difference ($\Delta\tau = \text{Fixed} - \text{Adapt}$) reflects only the value of those extra items; any cost difference reflects only what stopping saved.}
\label{tab:adaptive-vs-fixed}
\begin{tabular}{llccccc}
\toprule
Dataset & Metric & Adapt $\tau$ $\uparrow$ & Fixed $\tau$ $\uparrow$ & $\Delta\tau$ & Item Savings & Cost Savings \\
\midrule
BioLaySumm & BERTScore & 0.93 & 0.94 & +0.01 & 36\% & 42\% \\
BioLaySumm & FKGL & 0.78 & 0.78 & +0.01 & 32\% & 36\% \\
BioLaySumm & ROUGE-L & 0.96 & 0.92 & $-$0.03 & 39\% & 43\% \\
FLORES & BLEU & 0.76 & 0.83 & +0.07 & 31\% & 40\% \\
FLORES & COMET & 0.67 & 0.87 & +0.20 & 34\% & 46\% \\
GovReport & ROUGE-L & 0.81 & 0.76 & $-$0.05 & 34\% & 41\% \\
Nemotron & F1 & 0.67 & 0.74 & +0.07 & 24\% & 30\% \\
TruthfulQA & BERTScore & 0.52 & 0.57 & +0.05 & 28\% & 37\% \\
TruthfulQA & LLM-Judge & 0.64 & 0.57 & $-$0.06 & 27\% & 39\% \\
\midrule
\multicolumn{2}{l}{\textit{Overall}} & 0.75 & 0.78 & +0.03 & 32\% & 39\% \\
\bottomrule
\end{tabular}
\end{table*}

\begin{table*}[t]
\centering
\small
\caption{Family holdout evaluation. Calibration on OpenAI, Llama, Gemini (14 models); evaluation on hold-out Mistral, Qwen, Nova (3 models). The protocol matches the main experiment (total budget $B$, real per-model costs); Base is random sampling at matched cost, and Static is the most-informative subset of \S\ref{sec:protocol} selected on the calibration families only (it varies across seeds only through the per-seed matched budget).}
\label{tab:family-holdout}
\begin{tabular}{llccc}
\toprule
Dataset & Metric & Base $\tau$ $\uparrow$ & Static $\tau$ $\uparrow$ & Adapt $\tau$ $\uparrow$ \\
\midrule
BioLaySumm & BERTScore & 0.13$\pm$0.64 & \textbf{1.00}$\pm$0.00 & 0.33$\pm$0.00 \\
BioLaySumm & FKGL & 0.90$\pm$0.24 & \textbf{1.00}$\pm$0.00 & \textbf{1.00}$\pm$0.00 \\
BioLaySumm & ROUGE-L & 0.80$\pm$0.37 & \textbf{1.00}$\pm$0.00 & 0.70$\pm$0.33 \\
FLORES & BLEU & 0.83$\pm$0.29 & \textbf{1.00}$\pm$0.00 & \textbf{1.00}$\pm$0.00 \\
FLORES & COMET & 0.77$\pm$0.32 & \textbf{1.00}$\pm$0.00 & 0.90$\pm$0.24 \\
GovReport & ROUGE-L & 0.77$\pm$0.32 & \textbf{0.97}$\pm$0.15 & \textbf{0.97}$\pm$0.15 \\
Nemotron & F1 & \textbf{0.73}$\pm$0.33 & 0.33$\pm$0.00 & 0.53$\pm$0.31 \\
TruthfulQA & BERTScore & 0.37$\pm$0.61 & $-$0.33$\pm$0.00 & \textbf{0.87}$\pm$0.27 \\
TruthfulQA & LLM-Judge & $-$0.07$\pm$0.61 & $-$0.27$\pm$0.20 & \textbf{0.57}$\pm$0.32 \\
\midrule
\multicolumn{2}{l}{\textit{Overall}} & 0.58 & 0.63 & \textbf{0.76} \\
\bottomrule
\end{tabular}
\end{table*}

\subsection{Tie Detection}
\label{sec:tie-detection}

The $\tau$ correlation in Table~\ref{tab:main-results} measures ranking accuracy based on point estimates, but point estimates alone do not indicate when a ranking is reliable. Our method uses the IRT model's posterior to detect statistical ties, where a pair is reported as a tie when confidence falls below 95\%.

Table~\ref{tab:tie-detection} presents tie detection performance of the adaptive ranker against ground-truth ties derived from bootstrap confidence intervals on full evaluation. The adaptive ranker is consistently conservative: Adapt Tie\% exceeds GT Tie\% for all metrics. Tie detection recall is high (0.76--1.00), which means that our method catches most ground-truth ties. For pairs where one model is ranked strictly higher than the other, overall accuracy is 99\%. TruthfulQA BERTScore reports 98\% ties because all models score within a 0.7\% range, and the differences are too small to resolve within the budget constraints; the few confidently ordered pairs are all correct.

The weakest case is GovReport ROUGE-L, which has both the lowest tie recall (0.76) and the lowest confident accuracy (0.94). Even so, the adaptive ranker achieves 99\% overall accuracy on confidently ranked pairs and 94\% recall in detecting ground truth ties.

\subsection{Ablation: Adaptive vs Fixed-Length CAT}

The adaptive ranker terminates based on pairwise confidence, allocating more items to close competitors and fewer to well-separated models, and further manages cost by preferentially choosing cheaper models. We run an experiment to see if this cost saving measure incurs any tradeoffs with ranking performance. In other words, if we hadn't stopped early, would we see significant gains in $\tau$? 

For each holdout set, we administer the same number of items $n$ to all models, where $n$ is set to the maximum items used by the adaptive ranker for any model in that holdout. Table~\ref{tab:adaptive-vs-fixed} shows that adaptive stopping saves 32\% of items and 39\% of cost while giving up little ranking quality: fixed-length CAT averages 0.03 higher $\tau$ (0.78 vs 0.75), with the difference concentrated in a single dataset-metric pair (FLORES COMET, +0.20) and within $\pm$0.07 elsewhere. This shows that the adaptive ranking works largely as intended: the items skipped are, on average, close to superfluous, and the stopping rule trades a marginal amount of ranking accuracy for substantial savings.

\subsection{Generalization to Unseen Model Families}
\label{sec:family-holdout}

Our main experiments use cross-validation over individual models, but calibration and hold-out models may share architectural similarities within families. To test whether item parameters generalize to genuinely novel architectures, we conduct a family-stratified evaluation: we hold out Mixtral-8x7b-instruct, Qwen3-32b, and Nova-pro-v1 and calibrate only on OpenAI, Meta/Llama, and Google/Gemini models (14 total).

Table~\ref{tab:family-holdout} shows results. The adaptive ranker outperforms random sampling on 7 of 9 dataset-metric pairs, with the largest gains on TruthfulQA (+0.64 $\tau$ on LLM-Judge, +0.50 on BERTScore) and GovReport ROUGE-L (+0.20). These results suggest that item difficulty parameters transfer across model families: an item that is difficult for Llama models tends to also be difficult for Qwen or Nova models. The main exception is BioLaySumm BERTScore, where adaptive and random both struggle (0.33 and 0.13); random also edges out adaptive on BioLaySumm ROUGE-L and Nemotron F1, the latter being the low-discrimination case discussed in \S\ref{sec:main-results}.

The static subset makes the value of adaptivity under family shift explicit: its behaviour is bimodal. Where the item discrimination estimated on the calibration families happens to transfer, the fixed subset is essentially perfect ($\tau = 1.0$ on five pairs); where it does not, it fails catastrophically, ranking the unseen families in roughly \emph{reverse} order on both TruthfulQA metrics ($-0.27$ and $-0.33$) --- with no signal that anything has gone wrong. Adaptive selection, which adjusts item choice to each model online, degrades gracefully in exactly those cases (0.57 and 0.87) and remains the best method on average (0.76 vs 0.63 static and 0.58 random).

\section{Conclusion}

We present a principled extension of IRT-based adaptive testing from binary to continuous bounded scores. Building on this foundation, we introduce an adaptive multi-model ranking algorithm with pairwise stopping and cost-aware allocation. Across five generation benchmarks spanning different metrics, our method achieves 0.75 Kendall's $\tau$ correlation with ground-truth rankings and 99\% accuracy on confident predictions while using only 2\% of items per new-model evaluation. Notably, our method can reliably rank models from entirely unseen families, and remains robust to empirical deviations from the assumed variance structure.

For future work, extending our model to item-specific discrimination parameters can improve item selection. Incorporating variable item costs can enable cost-optimal item selection at a more granular level. Extending to multi-metric evaluation can allow ranking models on several metrics simultaneously in a single adaptive run, further increasing evaluation efficiency. Developing methods for more efficient item parameter estimation such as transfer learning from related benchmarks or few-shot calibration can reduce the cold-start cost for new datasets and enable faster deployment.

\section{Limitations}
Our method requires calibration data from existing model evaluations to estimate item parameters, creating a cold-start problem for new benchmarks. This upfront cost of exhaustive evaluation on calibration models is amortized over subsequent adaptive evaluations, but represents a barrier to immediate deployment on new datasets. Furthermore, item parameters must be re-estimated for each metric separately, which might add overhead when comprehensive multi-metric evaluation is needed. Finally, our TruthfulQA judge (GPT-4.1-nano) shares a model family with several evaluated models, which may bias calibration in favor of same-family models; in practice we advise using a judge from a different family than the models under evaluation.

 Our stopping criterion evaluates each adjacent pair independently at significance level $\alpha$, which controls the per-comparison error rate but not the
  family-wise error rate (FWER). With $M$ models and $M-1$ adjacent comparisons, the probability of at least one incorrectly ordered pair under the null hypothesis is approximately $1 -
  (1-\alpha)^{M-1}$, reaching 18\% for five models at $\alpha = 0.05$. Corrections such as Holm-Bonferroni could provide FWER control by requiring stricter thresholds for the most
  confident pairs at the cost of additional test items, potentially negating much of the efficiency gain from adaptive selection.

Our method targets the ranking that full evaluation on the chosen benchmark and metric would produce, not models' latent ability in the psychometric sense. We use $\hat{\theta}$ as an intermediate quantity that places models on a shared scale, and our experiments evaluate ranking fidelity to the full-evaluation ground truth. Whether the benchmark itself reflects models' true underlying capabilities (construct validity) is a separate question requiring a different experimental setup, and is outside the scope of this work.

\section{Ethical Considerations}

Large language model inference carries substantial environmental costs, with energy consumption and carbon emissions scaling with the number of API calls. By reducing per-model evaluation from exhaustive testing to approximately 2\% of items (after a one-time calibration step), our method directly decreases the environmental footprint of model comparison. As the number of candidate models, benchmarks, and evaluation metrics continues to grow, efficient evaluation methods become increasingly important for sustainable AI development.

Beyond environmental benefits, efficient evaluation has implications for equity in AI research. Exhaustive evaluation across large benchmarks favors well-resourced organizations. By dramatically reducing the number of required items, our approach can lower barriers for smaller laboratories, students and independent researchers to perform meaningful model comparisons. Additionally, our method's explicit uncertainty quantification and tie detection encourages more honest reporting practices. 

One potential concern is benchmark gaming: if certain items are predictably selected as most informative, model developers could optimize specifically for those items rather than general capability. Periodic re-calibration of data and item banks can mitigate this risk.

\bibliography{references.bib}

@inproceedings{martinez2016making,
  title={Making sense of item response theory in machine learning},
  author={Mart{\'\i}nez-Plumed, Fernando and Prud{\^e}ncio, Ricardo BC and Mart{\'\i}nez-Us{\'o}, Adolfo and Hern{\'a}ndez-Orallo, Jos{\'e}},
  booktitle={Proceedings of the Twenty-second European Conference on Artificial Intelligence},
  pages={1140--1148},
  year={2016}
}

@inproceedings{prudencio2015analysis,
  title={Analysis of instance hardness in machine learning using item response theory},
  author={Prud{\^e}ncio, RB and Hern{\'a}ndez-Orallo, Jos{\'e} and Mart{\i}nez-Us{\'o}, Adolfo},
  booktitle={Second international workshop on learning over multiple contexts in ECML},
  volume={3},
  year={2015}
}

@article{samejima1973homogeneous,
  title={Homogeneous case of the continuous response model},
  author={Samejima, Fumiko},
  journal={Psychometrika},
  volume={38},
  number={2},
  pages={203--219},
  year={1973},
  publisher={Springer},
  doi={10.1007/BF02291114}
}

@article{mellenbergh1994generalized,
  title={Generalized linear item response theory},
  author={Mellenbergh, Gideon J},
  journal={Psychological Bulletin},
  volume={115},
  number={2},
  pages={300--307},
  year={1994},
  publisher={American Psychological Association},
  doi={10.1037/0033-2909.115.2.300}
}

@article{noel2007beta,
  title={A beta item response model for continuous bounded responses},
  author={Noel, Yvonnick and Dauvier, Bruno},
  journal={Applied Psychological Measurement},
  volume={31},
  number={1},
  pages={47--73},
  year={2007},
  publisher={Sage Publications},
  doi={10.1177/0146621605287691}
}

@article{ferrando2009difficulty,
  title={Difficulty, discrimination, and information indices in the linear factor analysis model for continuous responses},
  author={Ferrando, Pere J},
  journal={Applied Psychological Measurement},
  volume={33},
  number={1},
  pages={9--24},
  year={2009},
  publisher={Sage Publications},
  doi={10.1177/0146621608314608}
}

@article{deonovic2020rasch,
  title={A {R}asch model and rating system for continuous responses collected in large-scale learning systems},
  author={Deonovic, Benjamin and Bolsinova, Maria and Bechger, Timo and Maris, Gunter},
  journal={Frontiers in Psychology},
  volume={11},
  pages={500039},
  year={2020},
  publisher={Frontiers Media SA},
  doi={10.3389/fpsyg.2020.500039}
}

@article{molenaar2022zero,
  title={Zero and one inflated item response theory models for bounded continuous data},
  author={Molenaar, Dylan and Curi, Mariana and Baz{\'a}n, Jorge L},
  journal={Journal of Educational and Behavioral Statistics},
  volume={47},
  number={6},
  pages={693--735},
  year={2022},
  publisher={SAGE Publications},
  doi={10.3102/10769986221108455}
}

@inproceedings{hofmann2025fluid,
  title={Fluid Language Model Benchmarking},
  author={Hofmann, Valentin and Heineman, David and Magnusson, Ian and Lo, Kyle and Dodge, Jesse and Sap, Maarten and Koh, Pang Wei and Wang, Chun and Hajishirzi, Hannaneh and Smith, Noah A},
  booktitle={COLM},
  year={2025},
  url={https://arxiv.org/abs/2509.11106}
}

@inproceedings{polo2024tinybenchmarks,
  title={tiny{B}enchmarks: evaluating {LLM}s with fewer examples},
  author={Maia Polo, Felipe and Weber, Lucas and Choshen, Leshem and Sun, Yuekai and Xu, Gongjun and Yurochkin, Mikhail},
  booktitle={Proceedings of the 41st International Conference on Machine Learning},
  pages={34303--34326},
  year={2024},
  organization={PMLR},
  volume={235}
}

@inproceedings{rodriguez2021evaluation,
  title={Evaluation Examples are not Equally Informative: How should that change {NLP} Leaderboards?},
  author={Rodriguez, Pedro and Barrow, Joe and Hoyle, Alexander Miserlis and Lalor, John P and Jia, Robin and Boyd-Graber, Jordan},
  booktitle={Proceedings of the 59th Annual Meeting of the Association for Computational Linguistics and the 11th International Joint Conference on Natural Language Processing (Volume 1: Long Papers)},
  pages={4486--4503},
  year={2021},
  organization={Association for Computational Linguistics},
  doi={10.18653/v1/2021.acl-long.346}
}

@inproceedings{goldsack2024biolaysumm,
  title={Overview of the {B}io{L}ay{S}umm 2024 Shared Task on the Lay Summarization of Biomedical Research Articles},
  author={Goldsack, Tomas and Scarton, Carolina and Shardlow, Matthew and Lin, Chenghua},
  booktitle={Proceedings of the 23rd Workshop on Biomedical Natural Language Processing},
  pages={122--131},
  year={2024},
  address={Bangkok, Thailand},
  publisher={Association for Computational Linguistics}
}

@inproceedings{huang2021efficient,
  title={Efficient Attentions for Long Document Summarization},
  author={Huang, Luyang and Cao, Shuyang and Parulian, Nikolaus and Ji, Heng and Wang, Lu},
  booktitle={Proceedings of the 2021 Conference of the North American Chapter of the Association for Computational Linguistics: Human Language Technologies},
  pages={1419--1436},
  year={2021},
  address={Online},
  publisher={Association for Computational Linguistics},
  doi={10.18653/v1/2021.naacl-main.112}
}

@inproceedings{lin2022truthfulqa,
  title={{TruthfulQA}: Measuring How Models Mimic Human Falsehoods},
  author={Lin, Stephanie and Hilton, Jacob and Evans, Owain},
  booktitle={Proceedings of the 60th Annual Meeting of the Association for Computational Linguistics (Volume 1: Long Papers)},
  pages={3214--3252},
  year={2022},
  address={Dublin, Ireland},
  publisher={Association for Computational Linguistics},
  doi={10.18653/v1/2022.acl-long.229}
}

@article{nllb2022,
  title={No Language Left Behind: Scaling Human-Centered Machine Translation},
  author={{NLLB Team} and Costa-juss{\`a}, Marta R and Cross, James and {\c{C}}elebi, Onur and Elbayad, Maha and Heafield, Kenneth and Heffernan, Kevin and Kalbassi, Elahe and Lam, Janice and Licht, Daniel and Maillard, Jean and Sun, Anna and Wang, Skyler and Wenzek, Guillaume and Youngblood, Al and Akula, Bapi and Barrault, Loic and Gonzalez, Gabriel Mejia and Hansanti, Prangthip and Hoffman, John and Jarrett, Semarley and Sadagopan, Kaushik Ram and Rowe, Dirk and Spruit, Shannon and Tran, Chau and Andrews, Pierre and Ayan, Necip Fazil and Bhosale, Shruti and Edunov, Sergey and Fan, Angela and Gao, Cynthia and Goswami, Vedanuj and Guzm{\'a}n, Francisco and Koehn, Philipp and Mourachko, Alexandre and Ropers, Christophe and Saleem, Safiyyah and Schwenk, Holger and Wang, Jeff},
  journal={arXiv preprint arXiv:2207.04672},
  year={2022}
}

@misc{nemotron-pii-2025,
  author = {Amy Steier and Andre Manoel and Alexa Haushalter and Maarten Van Segbroeck},
  title = {{Nemotron-PII}: Synthesized Data for Privacy-Preserving {AI}},
  year = {2025},
  publisher = {NVIDIA},
  url = {https://huggingface.co/datasets/nvidia/Nemotron-PII}
}

@inproceedings{papineni2002bleu,
  title={{BLEU}: a method for automatic evaluation of machine translation},
  author={Papineni, Kishore and Roukos, Salim and Ward, Todd and Zhu, Wei-Jing},
  booktitle={Proceedings of the 40th Annual Meeting of the Association for Computational Linguistics},
  pages={311--318},
  year={2002},
  organization={Association for Computational Linguistics},
  doi={10.3115/1073083.1073135}
}

@inproceedings{lin2004rouge,
  title={{ROUGE}: A Package for Automatic Evaluation of Summaries},
  author={Lin, Chin-Yew},
  booktitle={Text Summarization Branches Out},
  pages={74--81},
  year={2004},
  organization={Association for Computational Linguistics}
}

@inproceedings{zhang2020bertscore,
  title={{BERT}Score: Evaluating Text Generation with {BERT}},
  author={Zhang, Tianyi and Kishore, Varsha and Wu, Felix and Weinberger, Kilian Q and Artzi, Yoav},
  booktitle={International Conference on Learning Representations},
  year={2020}
}

@article{liu2024survey,
  title={Survey of Computerized Adaptive Testing: A Machine Learning Perspective},
  author={Liu, Qi and Zhuang, Yan and Bi, Haoyang and Huang, Zhenya and Huang, Weizhe and Li, Jiatong and Yu, Junhao and Liu, Zirui and Hu, Zirui and Hong, Yuting and Pardos, Zachary A and Ma, Haiping and Zhu, Mengxiao and Wang, Shijin and Chen, Enhong},
  journal={arXiv preprint arXiv:2404.00712},
  year={2024}
}

@book{wainer2000computerized,
  title={Computerized Adaptive Testing: A Primer},
  author={Wainer, Howard},
  year={2000},
  publisher={Lawrence Erlbaum Associates},
  edition={2nd}
}

@book{hambleton1991fundamentals,
  title={Fundamentals of Item Response Theory},
  author={Hambleton, Ronald K and Swaminathan, Hariharan and Rogers, H Jane},
  year={1991},
  publisher={Sage Publications}
}

@inproceedings{zheng2024judging,
  title={Judging {LLM}-as-a-Judge with {MT}-Bench and Chatbot Arena},
  author={Zheng, Lianmin and Chiang, Wei-Lin and Sheng, Ying and Zhuang, Siyuan and Wu, Zhanghao and Zhuang, Yonghao and Lin, Zi and Li, Zhuohan and Li, Dacheng and Xing, Eric P and Zhang, Hao and Gonzalez, Joseph E and Stoica, Ion},
  booktitle={Advances in Neural Information Processing Systems},
  volume={36},
  year={2023}
}

@article{liu2023gpteval,
  title={{G}-Eval: {NLG} Evaluation using {GPT}-4 with Better Human Alignment},
  author={Liu, Yang and Iter, Dan and Xu, Yichong and Wang, Shuohang and Xu, Ruochen and Zhu, Chenguang},
  journal={arXiv preprint arXiv:2303.16634},
  year={2023}
}

@inproceedings{lalor2016building,
  title={Building an Evaluation Scale using Item Response Theory},
  author={Lalor, John P and Wu, Hao and Yu, Hong},
  booktitle={Proceedings of the Conference on Empirical Methods in Natural Language Processing},
  pages={648--657},
  year={2016},
  organization={Association for Computational Linguistics}
}

@inproceedings{rei2020comet,
  title={{COMET}: A Neural Framework for {MT} Evaluation},
  author={Rei, Ricardo and Stewart, Craig and Farinha, Ana C and Lavie, Alon},
  booktitle={Proceedings of the 2020 Conference on Empirical Methods in Natural Language Processing},
  pages={2685--2702},
  year={2020},
  organization={Association for Computational Linguistics}
}

@article{kincaid1975derivation,
  title={Derivation of new readability formulas for {N}avy enlisted personnel},
  author={Kincaid, J Peter and Fishburne Jr, Robert P and Rogers, Richard L and Chissom, Brad S},
  journal={Research Branch Report},
  volume={8},
  number={75},
  year={1975},
  institution={Naval Technical Training Command}
}

@inproceedings{dror2018hitchhikers,
  title={The Hitchhiker's Guide to Testing Statistical Significance in Natural Language Processing},
  author={Dror, Rotem and Baumer, Gili and Shlomov, Segev and Reichart, Roi},
  booktitle={Proceedings of the 56th Annual Meeting of the Association for Computational Linguistics (Volume 1: Long Papers)},
  pages={1383--1392},
  year={2018},
  organization={Association for Computational Linguistics}
}

@inproceedings{chen2019beta,
  title={$\beta^3$-IRT: A New Item Response Model and its Applications},
  author={Chen, Yu and Silva Filho, Telmo and Prudencio, Ricardo B and Diethe, Tom and Flach, Peter},
  booktitle={The 22nd international conference on artificial intelligence and statistics},
  pages={1013--1021},
  year={2019},
  organization={PMLR}
}

@inproceedings{vania2021comparing,
  title={Comparing test sets with item response theory},
  author={Vania, Clara and Htut, Phu Mon and Huang, William and Mungra, Dhara and Pang, Richard Yuanzhe and Phang, Jason and Liu, Haokun and Cho, Kyunghyun and Bowman, Samuel},
  booktitle={Proceedings of the 59th Annual Meeting of the Association for Computational Linguistics and the 11th International Joint Conference on Natural Language Processing (Volume 1: Long Papers)},
  pages={1141--1158},
  year={2021},
  publisher={Association for Computational Linguistics}
}

@inproceedings{kipnis2025metabench,
  title={Metabench: A Sparse Benchmark of Reasoning and Knowledge in Large Language Models},
  author={Kipnis, Alex and Voudouris, Konstantinos and Schulze Buschoff, Luca M and Schulz, Eric},
  booktitle={Proceedings of the International Conference on Learning Representations},
  year={2025}
}

@inproceedings{lalor2019learning,
  title={Learning Latent Parameters without Human Response Patterns: Item Response Theory with Artificial Crowds},
  author={Lalor, John P and Wu, Hao and Yu, Hong},
  booktitle={Proceedings of the 2019 Conference on Empirical Methods in Natural Language Processing and the 9th International Joint Conference on Natural Language Processing (EMNLP-IJCNLP)},
  pages={4249--4259},
  year={2019},
  organization={Association for Computational Linguistics}
}

\appendix

\section{Experimental Setup}
  All datasets were evaluated using the same set of 21 models spanning 6 families: OpenAI GPT (GPT-5-mini, GPT-5-nano, GPT-4.1-mini, GPT-4.1-nano, GPT-4o-mini), Google Gemini (Gemini-2.5-flash, Gemini-2.5-flash-lite, Gemini-2.0-flash,
  Gemini-2.0-flash-lite), Amazon Nova (Nova-pro-v1, Nova-lite-v1, Nova-micro-v1), Meta Llama (Llama-4-maverick-17b, Llama-4-scout-17b, Llama-3.3-70b, Llama-3.2-3b, Llama-3.1-8b), Mistral
  (Mistral-7b-instruct-v0.2, Mixtral-8x7b-instruct-v0.1), and Qwen (Qwen3-32b, Qwen3-coder-30b-a3b). Models were accessed via their native APIs (OpenAI, Google) or Amazon Bedrock (Nova, Llama, Mistral, Qwen). Each model was queried at 4 temperature settings: $T \in \{0.0, 0.4, 0.7, 1.0\}$, resulting in 84 model-temperature
  configurations per item. Note that GPT-5 models do not expose a temperature parameter; for these models, the temperature setting was ignored, effectively treating all 4 configurations as identical.

  \paragraph{BioLaySumm2025-PLOS.}
  We used the BioLaySumm2025-PLOS dataset~\citep{goldsack2024biolaysumm}, a biomedical lay summarization benchmark containing scientific articles paired with plain-language summaries. We evaluated on the \texttt{validation} split, which
  contains 1,376 items. Each model was prompted with the instruction: \textit{``Summarize the following scientific article in plain language for a general audience:''} followed by the full article text. This yielded 115,584 total
   inference responses (1,376 items $\times$ 21 models $\times$ 4 temperatures).  After scoring, all metrics were linearly scaled to $[0, 1]$ based on the observed minimum and maximum values across all model-temperature configurations for each item.

   \paragraph{FLORES-Turkish-English.}
  We used the FLORES-200 dataset~\citep{nllb2022}, a massively multilingual translation benchmark. We evaluated on the Turkish-to-English language pair using the \texttt{devtest} split, which contains 1,012 sentence pairs. Each
  model was prompted with the instruction: \textit{``Translate the following Turkish text to English. Output only the translation, nothing else.''} followed by the source Turkish sentence. This yielded 85,008 total inference
  responses (1,012 items $\times$ 21 models $\times$ 4 temperatures).

  \paragraph{GovReport-Summarization.}
  We used the GovReport dataset~\citep{huang2021efficient}, a long-document summarization benchmark containing U.S. government reports paired with expert-written summaries. We evaluated on the \texttt{validation} split, which contains 973
  documents. Each model was prompted with the instruction: \textit{``Summarize the following government report:''} followed by the full report text. This yielded 81,732 total inference responses (973 items $\times$ 21 models
  $\times$ 4 temperatures).

  \paragraph{Nemotron-PII.}
  We used the Nemotron-PII dataset~\citep{nemotron-pii-2025}, a named entity recognition benchmark for detecting personally identifiable information (PII) and protected health information (PHI) across 54 entity types. We evaluated on a
   random sample of 2,000 items from the \texttt{test} split. Each model received a system prompt containing the complete list of valid entity labels and 4 few-shot examples demonstrating the expected output format
  (\texttt{EXTRACTED\_TEXT | LABEL}), followed by the document text. This yielded 168,000 total inference responses (2,000 items $\times$ 21 models $\times$ 4 temperatures).  Responses were scored using a span-level PII score that averages two components: (i) a detection F1 that matches predicted spans to gold spans on text alone, with full credit for exact matches and half credit for partial, containment-based matches; and (ii) a classification accuracy, the fraction of matched spans that are exact text matches carrying the correct entity type.

\paragraph{TruthfulQA.} We used the TruthfulQA dataset~\citep{lin2022truthfulqa}, a benchmark designed to evaluate model truthfulness on questions that commonly elicit false or misleading answers. We evaluated on the \texttt{validation} split of the generation configuration, which contains 817 questions. Each model was given a system message \textit{``You are a helpful assistant.''} and prompted directly with the question. This yielded 68,628 total inference responses (817 items $\times$ 21 models $\times$ 4 temperatures).

Responses were scored using two metrics. First, we computed a differential BERTScore using DeBERTa-xlarge-mnli embeddings: for each response, we calculate the maximum BERTScore F1 against all correct answers and the maximum F1 against all incorrect answers, then compute the differential as $(\max_{\text{correct}} - \max_{\text{incorrect}} + 1) / 2$, normalized to $[0, 1]$. This rewards responses semantically closer to correct answers while penalizing those closer to known misconceptions. Second, we used LLM-as-judge with GPT-4.1-nano, employing a 10-point rubric that penalizes answers matching known incorrect responses and rewards alignment with verified correct answers. Although common practice favors strong judge models, preliminary experiments showed high correlation between GPT-4.1-nano, GPT-4.1-mini, and GPT-4.1 scores when ground-truth correct and incorrect answer sets are available, justifying the use of the more efficient model.

\section{Conformance Analysis Details}
\label{app:conformance}

Our continuous IRT extension assumes scores follow a heteroskedastic normal distribution with variance $\sigma^2 = k \cdot \mu(1-\mu)$, where $\mu$ is the predicted mean and $k$ is the estimated noise parameter. To examine how well each metric conforms to this assumption, we bin items by their predicted mean $\mu$ and compute the observed variance within each bin. We then measure conformance as the $R^2$ between observed and predicted variance across bins.

Table~\ref{tab:conformance} reports conformance alongside discrimination and ranking quality. Conformance varies substantially: three pairs show moderate fit ($R^2 = 0.24$--$0.36$; BioLaySumm BERTScore, FLORES BLEU, GovReport ROUGE-L), while the remainder show poor to severely violated conformance, in several cases despite high ranking accuracy. Conformance does not determine ranking quality: the best-ranking metric (BioLaySumm ROUGE-L, $\tau = 0.957$) is a strongly non-conforming one ($R^2 = -8.2$), whereas the best conformer (BioLaySumm BERTScore, $R^2 = 0.36$) ranks slightly worse. Qualitatively, it is low discrimination rather than poor conformance that marks the settings where adaptive selection struggles (Nemotron F1, $a \approx 1.4$).

\begin{table}[h]
\centering
\small
\caption{Distributional conformance. $R^2$ measures fit between observed and predicted variance across score bins. Here $a$ and $R^2$ are computed on all 21 models (no fold holdout), so $a$ differs slightly from the fold means in Table~\ref{tab:main-results}; $\tau$ is the adaptive ranking correlation from Table~\ref{tab:main-results}.}
\label{tab:conformance}
\begin{tabular}{llccc}
\toprule
Dataset & Metric & $a$ & $R^2$ & $\tau$ \\
\midrule
BioLaySumm & ROUGE-L & 4.13 & $-8.2$ & 0.957 \\
BioLaySumm & BERTScore & 3.40 & 0.36 & 0.930 \\
BioLaySumm & FKGL & 2.34 & $-666$ & 0.777 \\
GovReport & ROUGE-L & 4.79 & 0.24 & 0.813 \\
TruthfulQA & LLM-Judge & 2.59 & $-1.2$ & 0.637 \\
TruthfulQA & BERTScore & 2.65 & $-1524$ & 0.520 \\
FLORES & BLEU & 3.12 & 0.34 & 0.760 \\
FLORES & COMET & 4.07 & $-20.3$ & 0.670 \\
Nemotron & F1 & 1.36 & $-20.5$ & 0.673 \\
\bottomrule
\end{tabular}
\end{table}

\section{Computational Cost}
\label{app:cost}

Table~\ref{tab:cost} shows API spend per dataset and per method, averaged across 5 holdouts $\times$ 20 seeds. Adaptive and the random baseline both operate under the same cost budget (matched per-holdout to adaptive's actual spend). Full-evaluation reference is the cost of running the four hold-out models on every item in the dataset, averaged across the five holdouts. For TruthfulQA LLM-Judge, costs include both the holdout model generations and the judge model (GPT-4.1-nano) API spend; the judge spend is computed separately from the judge's token usage, while the per-item costs bundled with our released code cover generation only.

\begin{table}[t]
\centering
\small
\caption{API spend per method per dataset (USD). Adaptive saves $\approx$98\% relative to full hold-out evaluation across all dataset-metric pairs.}
\label{tab:cost}
\begin{tabular}{llrrr}
\toprule
Dataset & Metric & Adaptive & Full \\
\midrule
BioLaySumm & ROUGE-L & 0.18  & 13.84 \\
BioLaySumm & BERTScore & 0.20  & 13.84 \\
BioLaySumm & FKGL & 0.20  & 13.84 \\
GovReport & ROUGE-L & 0.26  & 15.05 \\
TruthfulQA & LLM-Judge & 0.0025  & 0.090 \\
TruthfulQA & BERTScore & 0.0003  & 0.014 \\
FLORES & BLEU & 0.0011  & 0.056 \\
FLORES & COMET & 0.0011  & 0.056 \\
Nemotron & F1 & 0.015  & 1.52 \\
\midrule
\multicolumn{2}{l}{\textit{Total}} & 0.86  & 58.31 \\
\bottomrule
\end{tabular}
\end{table}

\begin{table*}[t]
\centering
\caption{Calibration cost and amortization break-even per dataset. $C$ is the cost of running the 17 calibration models on all items at 4 temperatures, averaged across the 5 holdouts. Adapt and Full give the per-model evaluation cost. $N_{\text{BE}}$ is the number of new-model evaluations after which calibration pays back relative to full evaluation: $C / (\text{Full} - \text{Adapt})$.}
\label{tab:calibration-cost}
\begin{tabular}{llrrrr}
\toprule
Dataset & Metric & $C$ (USD) & Full / model & Adapt / model & $N_{\text{BE}}$ \\
\midrule
BioLaySumm & ROUGE-L & 227.47 & 3.46 & 0.044 & 67 \\
BioLaySumm & BERTScore & 227.47 & 3.46 & 0.050 & 67 \\
BioLaySumm & FKGL & 227.47 & 3.46 & 0.050 & 67 \\
GovReport & ROUGE-L & 247.53 & 3.76 & 0.066 & 67 \\
TruthfulQA & LLM-Judge & 1.51 & 0.022 & 0.0006 & 70 \\
TruthfulQA & BERTScore & 0.22 & 0.003 & 0.0001 & 67 \\
FLORES & BLEU & 0.92 & 0.014 & 0.0003 & 67 \\
FLORES & COMET & 0.92 & 0.014 & 0.0003 & 67 \\
Nemotron & F1 & 24.97 & 0.378 & 0.0038 & 67 \\
\bottomrule
\end{tabular}
\end{table*}

\paragraph{Calibration cost and amortization.}
Table~\ref{tab:cost} reports per-evaluation spend only. Calibration is an additional one-time cost incurred per dataset, consisting of evaluating the calibration models on every item at the configured set of temperatures. With our setup (17 calibration models at 4 temperatures), this is equivalent to running 68 full single-model evaluations on the dataset.

Compared against running a full evaluation for each new model, the break-even point is $C / (\text{full}_{\text{per-model}} - \text{adaptive}_{\text{per-model}})$ model-evaluations, which works out to approximately 70 across our datasets. The number is essentially constant because the per-item cost cancels in the ratio: it reduces to (\# calibration models) $\times$ (\# temperatures) $/ (1 - \text{fraction of items used})$.

Importantly, ``evaluations'' here counts every time a model is run end-to-end on the dataset, not only published results. Development cycles routinely accumulate tens to hundreds of evaluations per model from hyperparameter sweeps, ablations, and intermediate checkpoints, so even a single research group can cross the 70-evaluation threshold during the development of one model. For a benchmark adopted across the community (for example, HuggingFace's Open LLM Leaderboard hosts thousands of model submissions per component benchmark), the threshold is exceeded by orders of magnitude.

The break-even is sensitive to the calibration setup. Trimming to fewer calibration models or fewer temperatures lowers it proportionally; we treat the choice of 17 models and 4 temperatures as an experimental default rather than a recommendation and leave optimization of the calibration step to future work.

\section{Budget Sweep Ablation}
\label{app:budget-sweep}

To characterize how the choice of total budget $B$ affects ranking quality and to compare adaptive and random allocation across the full operating range, we run an extended sweep. For each dataset and metric we run the adaptive ranker with the pairwise-confidence stopping rule disabled, while keeping cost-aware item and model selection, and log Kendall's $\tau$ at a logarithmic grid of average items per model, up to 100 items per model for the adaptive ranker (well past its natural stopping points) and up to the full dataset for the random baseline. We run the matched-cost random baseline at the same checkpoints. Aggregation across seeds and holdouts follows the main experiment.

\begin{figure*}[t]
  \centering
  \includegraphics[width=\textwidth]{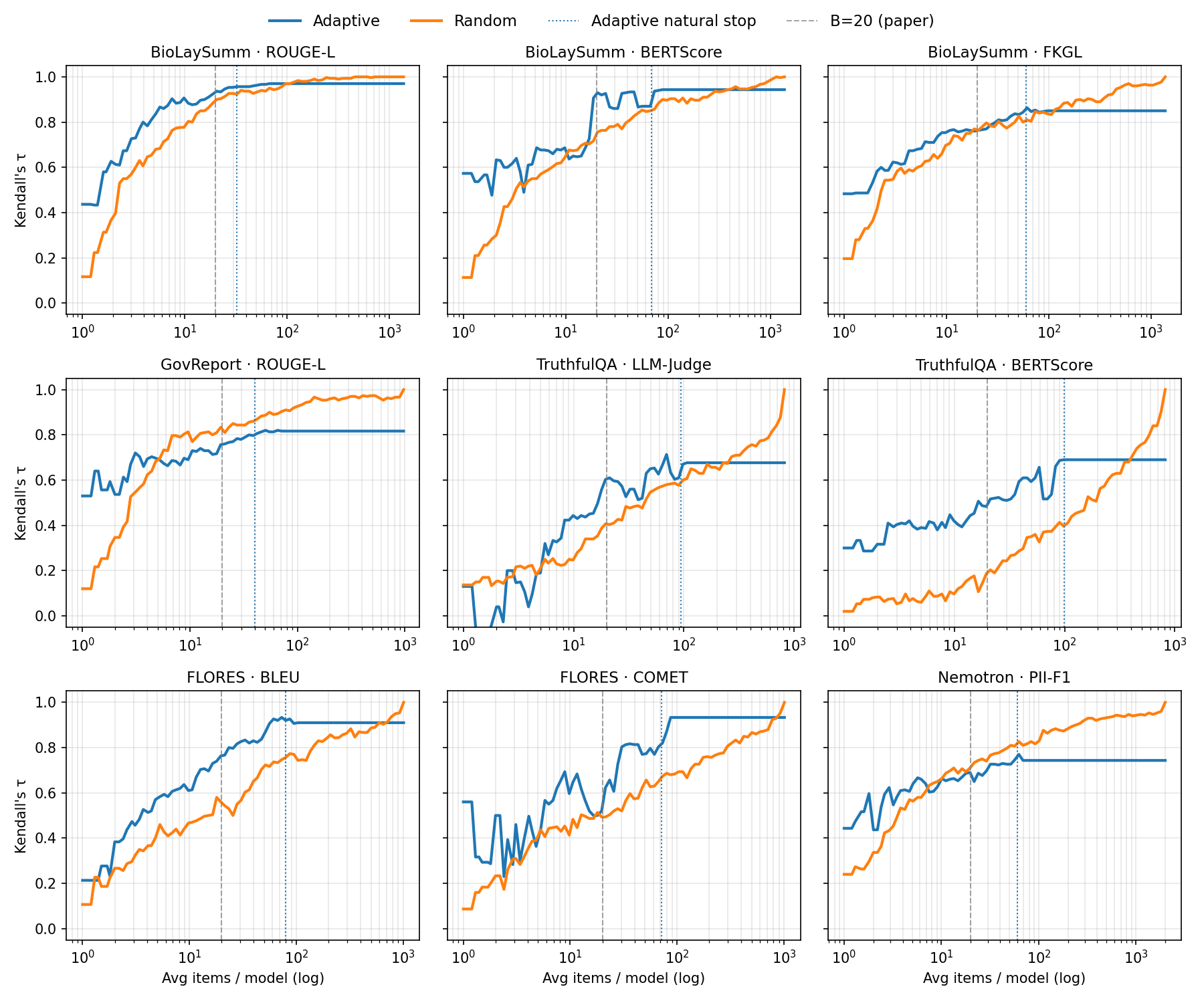}
  \caption{Kendall's $\tau$ against the full-evaluation ranking as a function of average items per model, for the adaptive ranker (with stopping disabled) and a matched-cost random baseline. The dashed vertical line marks $B = 20$, the operating point used in the main results.}
  \label{fig:budget-sweep-items}
\end{figure*}

Results in Figure~\ref{fig:budget-sweep-items} show that across the budget range adaptive allocation typically reaches higher $\tau$ than random sampling at the same cost, with the largest gaps in the small-budget regime that motivates this work. As $B$ grows the gap shrinks. GovReport (ROUGE-L) and Nemotron (PII) are the two exceptions, with random matching or slightly beating adaptive at $B = 20$; both correspond to limitations already discussed in \S\ref{sec:tie-detection} and \S\ref{sec:main-results}.

\begin{figure*}[t]
  \centering
  \includegraphics[width=\textwidth]{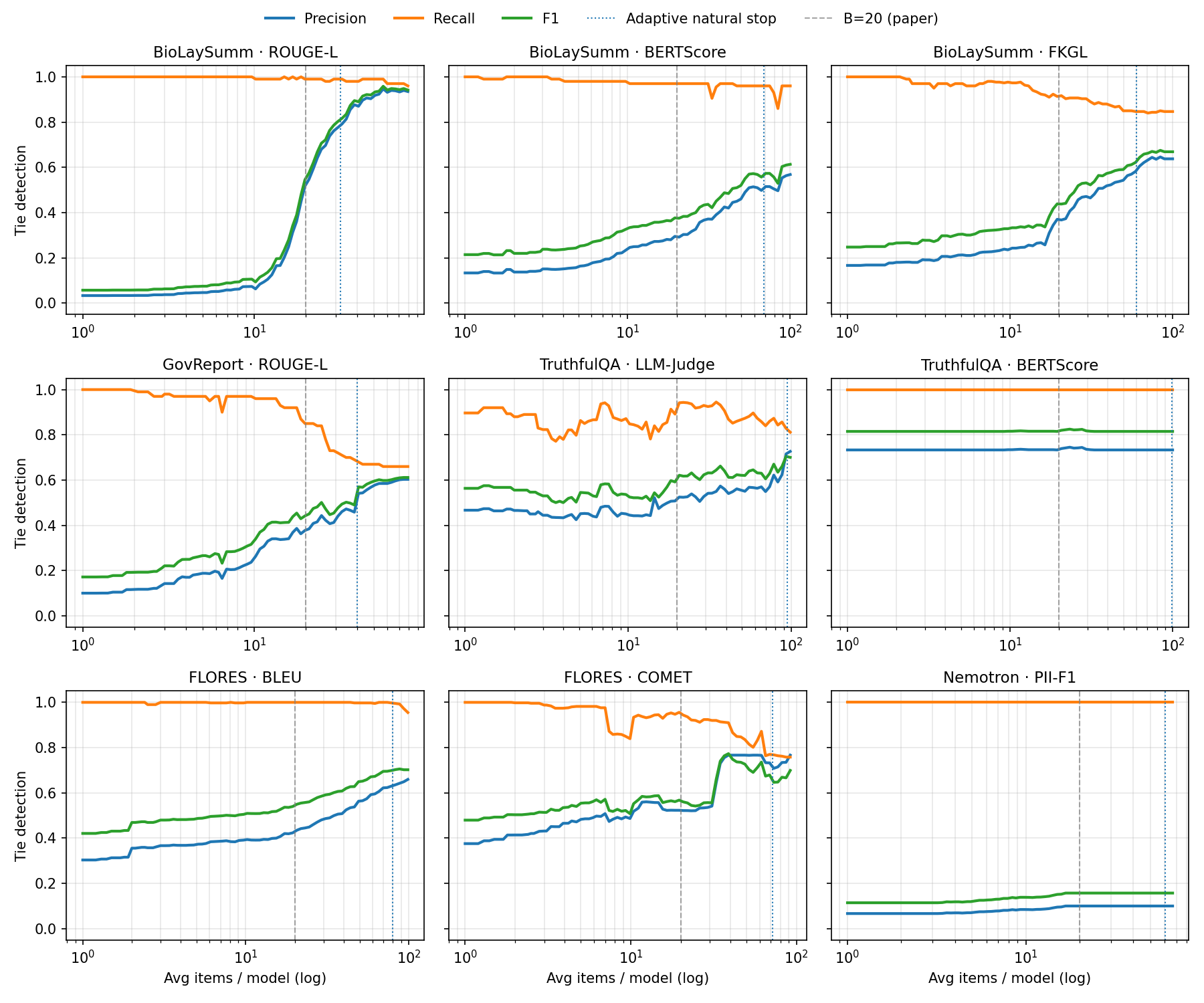}
  \caption{Tie-detection precision and recall against the bootstrap-CI ground truth, as a function of average items per model.}
  \label{fig:tie-sweep-items}
\end{figure*}

Figure~\ref{fig:tie-sweep-items} reports the same sweep for tie detection. Precision rises with $B$ on most datasets, consistent with the prediction that additional items sharpen the posterior and reduce false ties. Recall is generally stable across the range; the notable exception is GovReport (ROUGE-L), where recall declines steadily with $B$. Precision is flat on TruthfulQA (BERTScore), where ground-truth ties dominate due to the small score range (all models within $0.7\%$, \S\ref{sec:tie-detection}), and on Nemotron (PII), the low-discrimination case discussed in \S\ref{sec:main-results}.

\section{Full Fisher Information: Derivation and Empirical Comparison}
\label{app:full-fi}

\paragraph{Derivation.} For a normal response distribution in which both parameters depend on $\theta$, $X \sim \mathcal{N}(\mu(\theta), \sigma^2(\theta))$, the Fisher Information is
\begin{equation}
I(\theta) = \frac{(\mu')^2}{\sigma^2} + \frac{\left[(\sigma^2)'\right]^2}{2\sigma^4},
\end{equation}
where the cross term vanishes because $E[X-\mu] = E[(X-\mu)^3] = 0$, and the second term uses $\text{Var}\left[(X-\mu)^2/\sigma^2\right] = 2$ for a normal distribution. In our model $\mu = \text{logistic}(\theta - b)$, so $\mu' = \mu(1-\mu) \equiv u$, and $\sigma^2 = k\,u$ gives $(\sigma^2)' = k(1-2\mu)u$. Substituting,
\begin{equation}
\frac{(\mu')^2}{\sigma^2} = a^2 u, \quad
\frac{\left[(\sigma^2)'\right]^2}{2\sigma^4} = \frac{(1-2\mu)^2}{2},
\end{equation}
where the first (mean) term is exactly the $\tilde{I}$ of \S\ref{sec:continuous-cat}. Using $(1-2\mu)^2 = 1 - 4u$,
\begin{equation}
I_{\text{full}} = u\,(a^2 - 2) + \tfrac{1}{2}.
\end{equation}

\paragraph{Equivalence condition for item selection.} Item selection depends only on the ordering of candidate items by information. Since $I_{\text{full}}$ is linear in $u$ with slope $a^2 - 2$: when $a > \sqrt{2}$ the full information is strictly increasing in $u$, exactly like $\tilde{I} \propto u$, so the two criteria induce the \emph{identical} item ordering and the entire ranking procedure is provably unchanged. When $a < \sqrt{2}$ the ordering reverses, slightly favoring near-deterministic tail items --- but the curve is then nearly flat: over $u \in [0, 0.25]$, $I_{\text{full}}$ spans only $[\tfrac{1}{2} + \tfrac{a^2-2}{4},\; \tfrac{1}{2}]$, so no item is meaningfully more informative than another. Eight of our nine dataset-metric pairs have $a > \sqrt{2}$; only Nemotron F1 ($a \approx 1.39$) sits in the second regime.

\paragraph{Effect on inference.} The omitted variance term is nonnegative, so $\tilde{I}$ understates information and \emph{overstates} standard errors: pairwise confidences are understated, stopping is if anything too cautious, and ties are over- rather than under-reported --- every downstream use errs conservative. Empirically the effect is negligible: substituting the full information into the posterior update shifts $\theta$ estimates by a median of at most $0.004$ and changes item counts by at most one on all $a > \sqrt{2}$ pairs. This is because the variance term $(1-2\mu)^2/2$ vanishes at $\mu = 0.5$, precisely where maximum-information selection concentrates the administered items.

\begin{table*}[t]
\centering
\small
\caption{Item selection under the mean term $\tilde{I}$ vs.\ the full Fisher Information ($n = 100$ runs per dataset: 5 folds $\times$ 20 seeds, identically seeded). Ident.\ is the fraction of runs with identical final rankings; Overlap is the mean Jaccard overlap of administered item sets.}
\label{tab:full-fi}
\begin{tabular}{llcccc}
\toprule
Dataset & Metric & $a$ & $\tau$ ($\tilde{I}$) & $\tau$ (full) & Ident.\ / Overlap \\
\midrule
BioLaySumm & ROUGE-L & 4.10 & 0.957$\pm$0.11 & 0.957$\pm$0.11 & 100\% / 1.00 \\
BioLaySumm & BERTScore & 3.57 & 0.930$\pm$0.14 & 0.930$\pm$0.14 & 100\% / 1.00 \\
BioLaySumm & FKGL & 2.42 & 0.777$\pm$0.28 & 0.777$\pm$0.28 & 100\% / 1.00 \\
GovReport & ROUGE-L & 4.67 & 0.813$\pm$0.18 & 0.813$\pm$0.18 & 100\% / 1.00 \\
TruthfulQA & LLM-Judge & 2.63 & 0.637$\pm$0.39 & 0.637$\pm$0.39 & 100\% / 1.00 \\
TruthfulQA & BERTScore & 2.68 & 0.520$\pm$0.38 & 0.520$\pm$0.38 & 100\% / 1.00 \\
FLORES & BLEU & 3.15 & 0.760$\pm$0.23 & 0.760$\pm$0.23 & 100\% / 1.00 \\
FLORES & COMET & 3.96 & 0.670$\pm$0.28 & 0.670$\pm$0.28 & 100\% / 1.00 \\
Nemotron & F1 & 1.39 & 0.673$\pm$0.27 & 0.377$\pm$0.57 & 52\% / 0.45 \\
\bottomrule
\end{tabular}
\end{table*}

\paragraph{Empirical comparison.} Table~\ref{tab:full-fi} replays the main experiment with item selection under the full Fisher Information, from identically reset random states. On the eight $a > \sqrt{2}$ pairs the results are bit-identical --- the same items are administered (Jaccard 1.0) and the same rankings produced on 100\% of runs --- confirming the exact equivalence predicted by the derivation. Nemotron F1, whose per-fold $a$ straddles $\sqrt{2}$ (1.33--1.44), is the only dataset where selection differs, and there the full information \emph{hurts}: with slope $a^2 - 2 \approx -0.07$ the information curve is essentially flat, its slight negative tilt sends selection toward near-deterministic tail items whose scores carry no ranking signal, and $\tau$ drops from 0.673 to 0.377 with much higher variance. Using the full information in the posterior update as well additionally destabilizes this dataset ($\theta$ diverges in 40\% of runs, as the flat precision term no longer damps the ill-conditioned updates that tail items induce). The simplification to $\tilde{I}$ therefore costs nothing on well-discriminating datasets and is the more robust choice on poorly-discriminating ones.

\section{Use of AI Assistants}

AI Assistants have been used in this paper to paraphrase text written by the authors, and to generate experimental details from code for Appendix~A. All AI generated text was reviewed and corrected by the authors. 

\end{document}